\pdfoutput=1

\documentclass[11pt]{article}

\usepackage[final]{acl}

\usepackage{times}
\usepackage{latexsym}

\usepackage[T1]{fontenc}

\usepackage[utf8]{inputenc}

\usepackage{microtype}
\usepackage{amsmath}
\usepackage{amsfonts}
\usepackage{booktabs}
\usepackage{pdfpages}
\usepackage{multirow}
\usepackage{graphicx}
\newtheorem{definition}{Definition}
\title{Explaining the Effectiveness of Multi-Task Learning for Efficient Knowledge Extraction from Spine MRI Reports}

\author{Arijit Sehanobish\thanks{~~Corresponding Author},  {\bf McCullen Sandora,}  {\bf Nabila Abraham,} \\
{\bf Jayashri Pawar,}  {\bf Danielle Torres,}
{\bf Anasuya Das,} {\bf Murray Becker,} \\
{\bf Richard Herzog,} {\bf Benjamin Odry,} {\bf Ron Vianu} \\
Covera Health, New York City, New York \\
\texttt{\{arijit.sehanobish, mccullen.sandora, nabila.abraham,} \\
\texttt{jayashri.pawar, danielle.torres, anasuya.das, rherzog,} \\
\texttt{murray.becker, benjamin.odry, ron.vianu\}@coverahealth.com} \\
}

\begin{document}
\maketitle
\begin{abstract}
Pretrained Transformer based models finetuned on domain specific corpora have changed the landscape of NLP. However, training or fine-tuning these models for individual tasks can be time consuming and resource intensive. Thus, a lot of current research is focused on using transformers for multi-task learning~\citep{raffel2020exploring} and how to group the tasks to help a multi-task model to learn effective representations that can be shared across tasks~\citep{standley2020tasks, fifty2021efficiently}.  In this work, we show that a single multi-tasking model can match the performance of task specific models when the task specific models show similar representations across all of their hidden layers and their gradients are aligned, i.e. their gradients follow the same direction. We hypothesize that the above observations explain the effectiveness of multi-task learning. We validate our observations on our internal radiologist-annotated datasets on the cervical and lumbar spine. Our method is simple and intuitive, and can be used in a wide range of NLP problems. 
\end{abstract}
\begin{figure*}[ht]
	\centering
	\includegraphics[width=.9\textwidth]{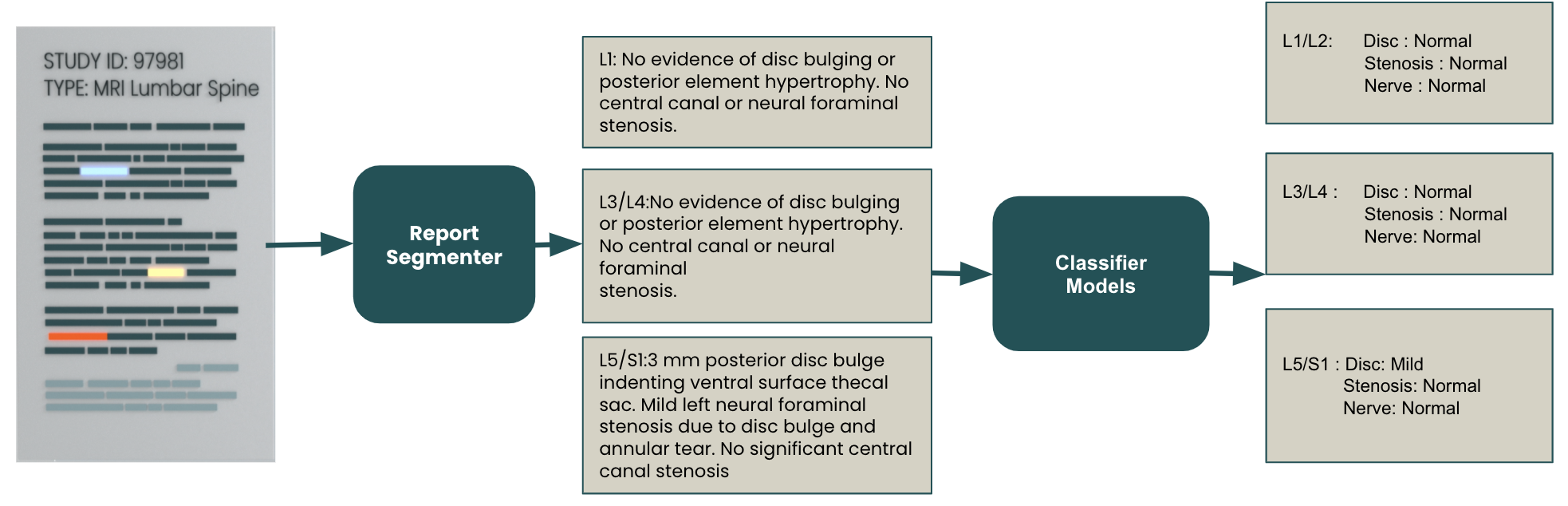}

	\caption{Figure showing how a report looks as it goes through our pipeline.}
	\label{Fig:lumbarworkflow}
\end{figure*}
\section{Introduction}

Since the seminal work by~\citep{vaswani2017attention}, Transformers have become the main architecture for almost all Natural Language Processing (NLP) tasks. Self-supervised pretraining of massive language models like BERT~\citep{devlin2019bert} and GPT~\citep{brown2020language} has allowed practitioners to use these large language models with little or no finetuning to various downstream tasks.  Multi-task learning (MTL) in NLP has been a very promising approach and has shown to lead to performance gains even over task specific fine-tuned models~\citep{Worsham2020, raffel2020exploring, aribandi2021ext5}. However, applying these large pre-trained Transformer models to downstream medical NLP tasks is quite difficult. Medical NLP has its unique challenges ranging from domain specific corpora, noisy annotation labels and scarcity of high quality labeled data. Despite these challenges, a number of researchers and practitioners have successfully finetuned these large language models for various medical NLP tasks. However, there is not much literature that uses multi-task learning in medical NLP to classify and extract diagnoses from clinical text~\citep{peng2020empirical, Crichton2017}. Moreover, there is almost no work in predicting spine pathologies from the radiologists' notes~\cite{asj}.

In this article, we are interested in extracting information from radiologists' notes on the cervical and the lumbar spine. In a given note, the radiologist discusses the specific, and often multiple pathologies, present in the medical images and grade their severity. Extracting relevant pathologies from these reports can facilitate the creation of structured databases that can be used for a number of downstream use-cases, such as cohort creation, quality assessment and outcome tracking. Single-task learning for information extraction in medical NLP has enjoyed much success in deep learning~\citep{Kanakarajan2021BioELECTRAPretrainedBT}. 

However, an ultimate NLP system for a complete understanding of the medical report must be able to perform many diverse information extraction and classification tasks simultaneously and efficiently. Such a system can be enabled by MTL, where one model shares weights across multiple tasks and makes multiple inferences in one forward pass. Such networks can not only be trained with limited resources, but are more scalable and deployable when compared to several single-task models. Moreover, the shared features within these MTL networks can induce more robust regularization and boost performance. Thus there is a lot of interest in the academic and industry research communities to understand when multi-task learning improves performance over single-tasking models~\citep{crawshaw2020multitask}, and how to group a diverse set of tasks to encourage the model to learn a representation that can be shared across tasks~\citep{standley2020tasks, fifty2021efficiently, bingel-sogaard-2017-identifying, zamir2020robust}. Some of the aforementioned works, most notably in~\citep{wass_mt}, define a notion of task similarity via the Wasserstein distance and show that a small Wasserstein distance between tasks aids in MTL. 

This work is an extension of our earlier work~\citep{cervicalabstract} where we used parameter efficient MTL models to extract information from cervical spine. In that work, we defined tasks as a conditional distribution over the classes, and we attributed our success of MTL to smaller Wasserstein distance between tasks. However, computing Wasserstein distance is expensive and suffers from the curse of dimensionality~\citep{cuturi2013sinkhorn}, which requires the number of samples to be significantly larger than the dimension of the representation ($768$ for many transformer models) in order for the distance to be accurately estimated. This prevents us from being able to estimate Wasserstein distance for some of our minority classes, which have about $200$ examples. Even for majority classes where we have about 5k samples, our work suffers from large error rates. Thus, to alleviate the above drawbacks, in this work, we sought to use methods that are applicable to small data regimes that lie in high dimensional space.

Inspired by the work of~\citep{yu2020gradient, graddrop} and~\citep{pmlr-v97-kornblith19a}, we hypothesize if the single-task models show similar representations across their hidden layers and the task specific gradients are \textit{aligned} (see Definition 1 in Section $4.2$), the multi-task model can match or outperform the task-specific, single-task models. We validate this hypothesis on two multi-task settings on our internal datasets: \textbf{(a)} Four of the most common pathologies in the cervical spine - central canal and foraminal stenosis, disc herniation and cord compression,
and \textbf{(b)} Three pathologies in the lumbar spine - central canal stenosis, disc herniation and nerve root impingement.

In this work, we (\textbf{a}) extend our novel pipeline to extract and predict the severity of various pathologies in the lumbar and cervical spine at \textit{each motion segment}, (\textbf{b}) compute Central Kernel Alignment (CKA) and show similarity between the transformer layers trained for individual tasks on a given dataset, (\textbf{c}) compute dot products between the gradients of the task specific loss functions with respect to various parameters and show that most of the gradients flow along a similar trajectory and (\textbf{d}) show how to leverage that information into a simple MTL framework allowing us to achieve significant model compression during deployment and also speed up our inference without sacrificing the accuracy of our predictions.

\section{Datasets}
We use an internal dataset consisting of radiologists' MRI reports on the cervical and the lumbar spine. Our dataset is heterogeneous and is diversely sampled from a large number of different radiology practices and medical institutions; the cervical MRI data consists of $1578$ reports from $97$ different radiology practices detailing various pathologies of the cervical spine and our lumbar MRI data contains $2004$ reports from $170$ different practices. 

We annotate the cervical reports with the $4$ following pathologies: spinal stenosis, disc herniation, cord compression, and neural foraminal stenosis, and the lumbar reports with the $3$ pathologies: disc herniation, spinal stenosis, and nerve impingement. Each of these pathologies is accompanied by an indication of severity. In the cervical reports, the three categories for the central canal stenosis are based on gradation; none/mild are not clinically significant, moderate and severe definitions involve cord compression or flattening. The moderate versus severe gradation refers to the varying degrees of cord involvement. For disc herniation and central canal stenosis, the categories are based on a continuous spectrum and it is a standard practice in radiology for any continuous spectrum to be bucketed in mild, moderate and severe discrete categories. Cord compression severity is binary: compression/signal change versus none. This is because both cord compression and signal change can cause symptoms, and are therefore clinically relevant. Foraminal stenosis is treated as a binary task as well: severe versus non-severe, as severe foraminal stenosis may indicate nerve impingement, which is clinically significant. Similar considerations are taken into account when annotating the lumbar reports. The splits and the details of each category can be found in Table~\ref{tab:Dataset-stats}. The data distribution is highly imbalanced, and about $25\%$ of these reports are OCR-ed, which leads to additional challenges stemming from bad OCR errors. 

\begin{table}[h]
\centering
\resizebox{\columnwidth}{!}
{
\begin{tabular}{@{}l l l l @{}}
\toprule
Dataset &
  Pathology &
  Training Label Distribution &
  Test Label Distribution \\ \midrule
\multirow{3}{*}{Lumbar} &
  Disc &
  \begin{tabular}[c]{@{}l@{}}None/Mild : 1885\\ Moderate : 1998\\ Severe : 456\end{tabular} &
  \begin{tabular}[c]{@{}l@{}}None/Mild : 1068\\ Moderate :1588\\ Severe :332\end{tabular} \\ \cmidrule(l){2-4} 
 &
  Stenosis &
  \begin{tabular}[c]{@{}l@{}}None/Mild : 3787\\ Moderate : 350\\ Severe : 202\end{tabular} &
  \begin{tabular}[c]{@{}l@{}}None/Mild : 2411\\ Moderate : 304\\ Severe : 273\end{tabular} \\ \cmidrule(l){2-4} 
 &
  Nerve &
  \begin{tabular}[c]{@{}l@{}}Normal : 3790\\ Abnormal : 549\end{tabular} &
  \begin{tabular}[c]{@{}l@{}}Normal : 2376\\ Abnormal : 612\end{tabular} \\ \midrule
\multirow{4}{*}{Cervical} &
  Disc &
  \begin{tabular}[c]{@{}l@{}}None/Mild : 2731\\ Moderate : 2699\\ Severe : 797\end{tabular} &
  \begin{tabular}[c]{@{}l@{}}None/Mild : 401\\ Moderate : 378\\ Severe : 101\end{tabular} \\ \cmidrule(l){2-4} 
 &
  Stenosis &
  \begin{tabular}[c]{@{}l@{}}None/Mild : 5488\\ Moderate : 561\\ Severe : 178\end{tabular} &
  \begin{tabular}[c]{@{}l@{}}None/Mild : 793\\ Moderate : 68\\ Severe : 19\end{tabular} \\ \cmidrule(l){2-4} 
 &
  Cord Compression &
  \begin{tabular}[c]{@{}l@{}}Normal : 5702\\ Abnormal : 525\end{tabular} &
  \begin{tabular}[c]{@{}l@{}}Normal : 806\\ Abnormal : 74\end{tabular} \\ \cmidrule(l){2-4} 
 &
  Neural Foraminal Stenosis &
  \begin{tabular}[c]{@{}l@{}}Normal : 5262\\ Abnormal : 965\end{tabular} &
  \begin{tabular}[c]{@{}l@{}}Normal : 789\\ Abnormal : 91\end{tabular} \\ \bottomrule
\end{tabular}
}
\caption{Table showing statistics of our datasets}
\label{tab:Dataset-stats}
\end{table}

%

For a given report, each task is to predict the severity of a pathology for each motion segment - the smallest physiological motion unit of the spinal cord~\cite{Swartz2005}. Breaking information down at the motion segment level in this way enables pathological findings to be correlated with clinical exam findings, and can inform future treatment interventions. 

Every report is tagged by annotators with labels for relevant pathologies and severities, along with span information indicating which part(s) of the report mentions each pathology.  For example, in a report for the lumbar spine, the sentence ``L1-L2: There is no disc herniation. No spinal canal or foraminal narrowing" would be given normal or $0$ class for each of the $3$ pathologies (central canal stenosis, disc herniation and nerve root impingement). Similarly in a cervical spine report, the sentence `` C2-3:  Normal; no disc herniation or bulge.  No central canal stenosis or neuroforaminal narrowing" would be given a normal or $0$ class for all the $4$ pathologies. An example of a full radiology report can be found in Appendix A.

\section{Workflow}
In this section, we will briefly describe our pipeline. The reports are first de-identified according to HIPAA regulations. Next, a Spacy~\citep{spacy} parser is used to break the report into sentences.

A BERT based NER model which we call the \emph{report segmenter} is then used to identify the motion segment(s) referenced in each sentence, and all the sentences containing a particular motion segment are concatenated together. This report segmenter has been shown to achieve an F1 score of $.9$ on our internal datasets, and the same model is common across both the lumbar and the cervical datasets. More details about the NER model and the hyperparameters used to train it can be found in Appendix B and C. All pathologies are predicted using the concatenated text for a particular motion segment. Finally, the severities for each pathology are modeled as multi-label classification problem, and a pre-trained transformer is finetuned using the text for each motion segment. 

For more details about our pipeline and data processing, please see Appendix B.
Figure~\ref{Fig:lumbarworkflow} breaks down how a report looks as it is processed through our spine pipeline.

\begin{figure*}[t]
\centering
\includegraphics[width=16cm]{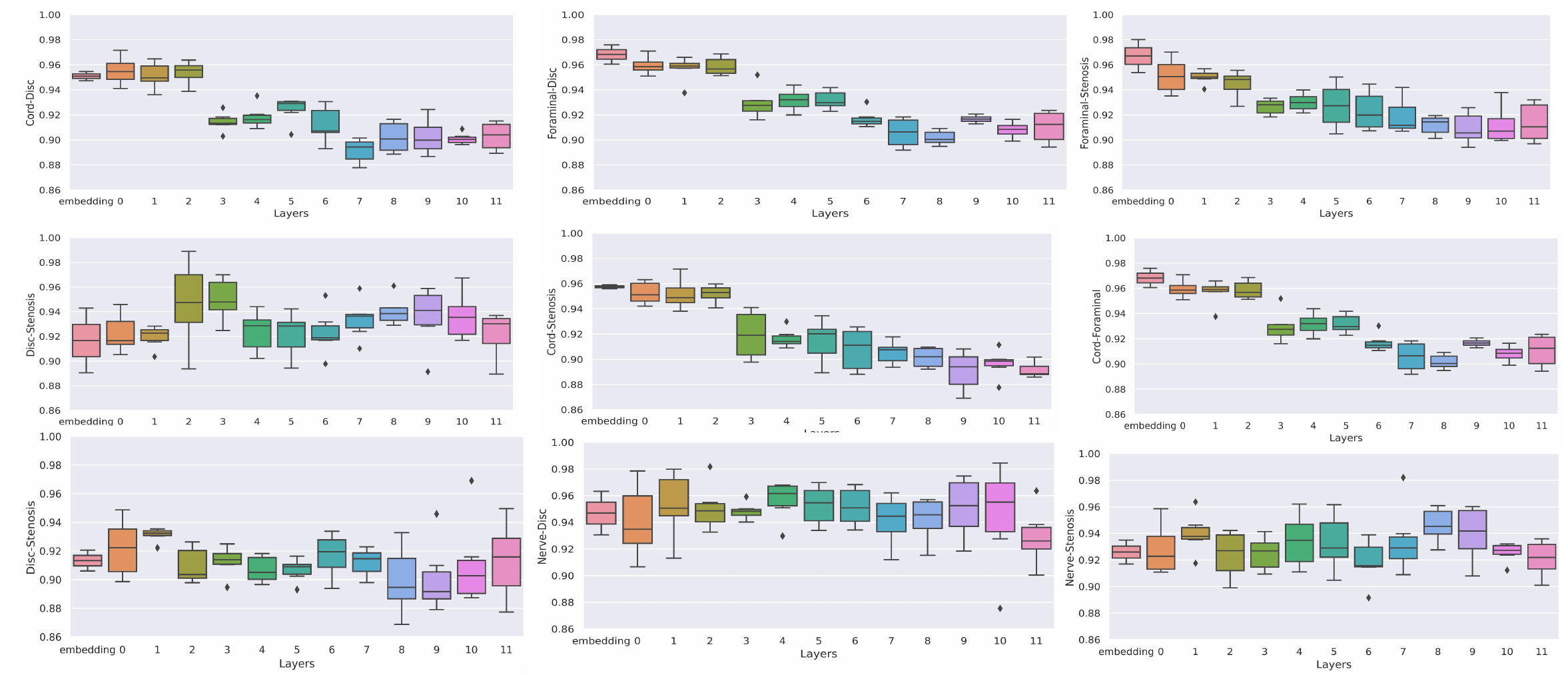}
\caption{CKA between activation matrices between different finetuned single-task models. The top 2 rows are single-task models trained to predict specific pathologies from cervical dataset and the bottom row for the lumbar dataset. The y-axis is chosen to be between the min and the max values, i.e. in the interval (.86, 1.0)}
\label{fig:cka}
\end{figure*} 

\section{Similarity of Representations between Task Specific Models}

In this section we will describe our methodology to understand the similarity between the representations of various single task models. For all the experiments in this section, we use the PubMedBERT~\cite{gu2020domain-specific} as the backbone. 

\subsection{Central Kernel Alignment}
We use the linear Central Kernel Alignment (CKA), introduced in~\citep{pmlr-v97-kornblith19a}. CKA is a scalar similarity index that can be used to compare representations within and across neural networks. (Linear) CKA can be defined by the following: Given $N$ examples and two activation outputs on these examples, $R_1 \in \mathbb{R}^{N \times d_1}$ and $R_2 \in \mathbb{R}^{N \times d_2}$,
\begin{equation}
\text{CKA}(R_1, R_2) = \frac{||R^{\top}_{1} R_2||_F}{||R^{\top}_{1} R_1||_F ||R^{\top}_{2} R_2||_F}
\end{equation}

where $|| \cdot ||_F$ is the Frobenius norm.

It is widely believed that similar representations lead to similar performances on downstream tasks~\citep{nguyen2021wide}. In this work, we compare the representations learned by various single tasking models. For two single task models trained on a specific part of a spine, the CKA between the matrix of activations for each layer of the corresponding models is computed. For illustration purposes, we collect all the CKA values for various activation matrices in a given layer and plot them in a box plot, as shown in figure~\ref{fig:cka}. We observe that for various tasks on both cervical and lumbar spine, all layers of the task specific models learn similar representations. 

Additional results on comparing models from the tasks from the lumbar dataset and the cervical dataset can be found in Appendix D.

However, the high value of CKA may also be attributed to the following factors : (i) larger and deeper networks converge to similar solutions~\citep{cca2018} and (ii) CKA values do not change drastically when models start from pretrained weights and are only trained for a few epochs~\citep{mtandcontinulalearning}. 

Thus in addition to the above analysis of the activations with the CKA, in the next subsection we look at the gradient level information to understand the trajectory of the task specific learned activations.

\subsection{Gradient Alignment}
There has been a lot of work in understanding the task specific gradients in the context of MTL. Given tasks $T_1, \cdots T_n$ (for example, they can be classification tasks), one can define $n$ loss functions $\mathcal{L}_{T_j}$ for each task $T_j$. In our work, all loss functions are cross-entropy losses. Then the task specific gradients are defined to be $\nabla_{\theta_j} \mathcal{L}_{T_j}$ where $\theta_j$ are the parameters of the task specific model. More specifically, it is shown in~\citep{chen2018gradnorm}, that MTL is competitive with single task learners when the norms of the task specific gradients have similar magnitudes. However in~\citep{yu2020gradient, rotograd}, the authors show that the direction of the gradient flow is more important than the magnitude for the success of MTL. More precisely, Theorem 1 in~\citep{yu2020gradient}, shows that the multitask objective converges to the optimum of one of the tasks or a sub-optimal minima in the presence of conflicting gradients. Furthermore, authors in~\citep{rotograd} use a synthetic toy example to show the difficulties of optimizing a multi-task loss in the presence of conflicting gradients. 

Inspired by the above works, we define the following:
\begin{definition}
Two gradient vectors $g_i$ and $g_j$ are \textit{aligned} if $g_i \cdot g_j > 0$, i.e. the vectors are pointing in the \textit{same direction}.
\end{definition}

\begin{figure*}[t]
\begin{center}
\includegraphics[width=16.2cm]{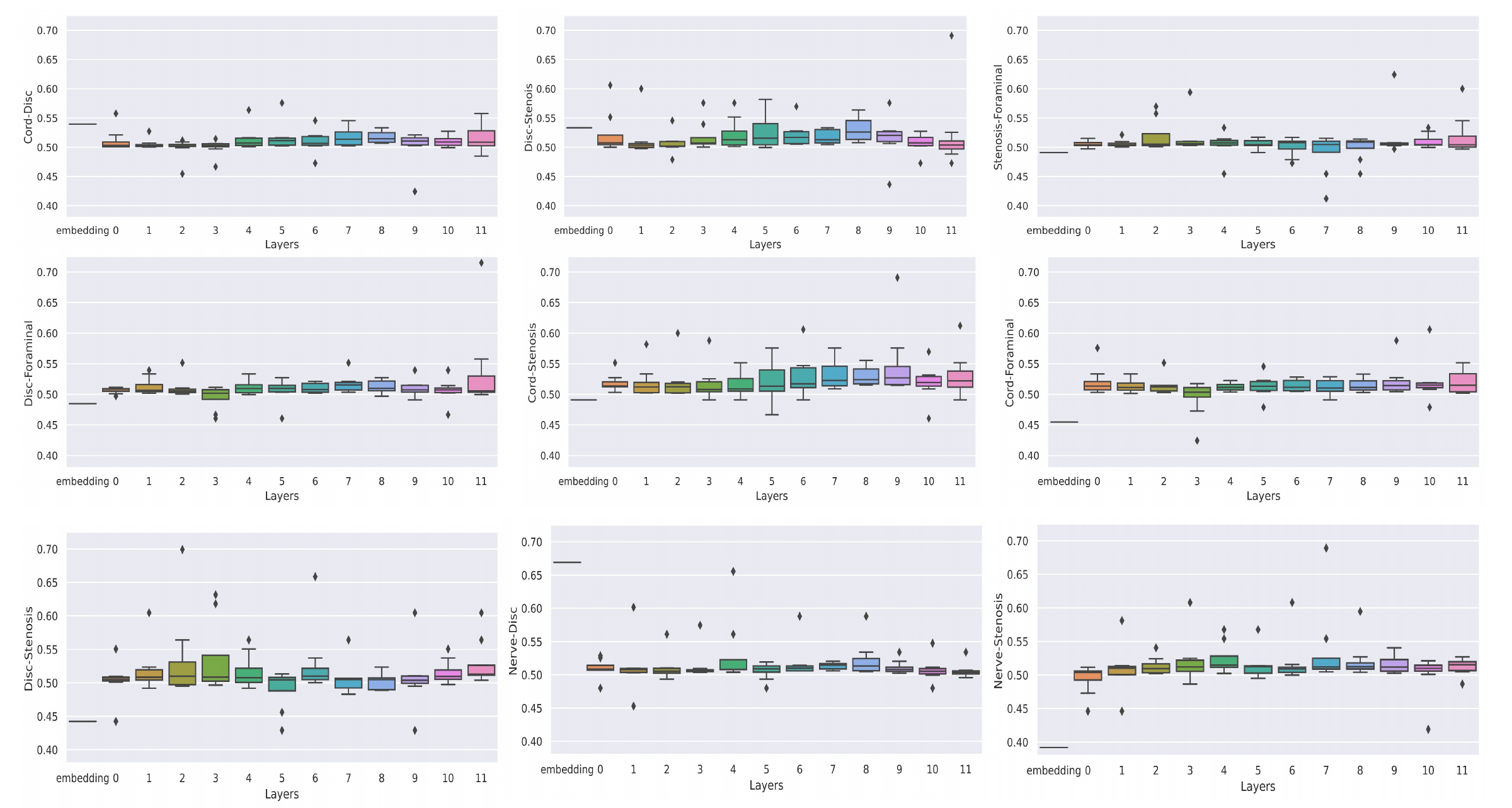}
\end{center}
\caption{Box Plot showing the proportion of aligned gradients between various task specific models, after training. The top 2 rows are single tasking models trained to predict specific pathologies from the cervical dataset and the bottom row for the lumbar dataset. The y-axis is chosen to be between the min and the max values, i.e. in the interval (.38, .725).}
\label{fig:grads}
\end{figure*}

To show that the gradients get more aligned as models are trained, we store the gradients for all the parameters for all the mini-batches after every epoch. We then compute the dot products between the corresponding gradients for two tasks. We observe that as the task specific models gets trained, an overwhelming proportion of these gradients are aligned (see Table~\ref{tab:align}). To illustrate our findings, we take the proportion of these aligned parameters in a given layer and plot them using a box plot in Figure~\ref{fig:grads}. Finally, we compute the proportion of weights across all layers for which the gradients are aligned which we call the Average Proportion of Aligned Gradients (APAG). 
\begin{equation}
    \text{APAG} = \frac1N_\text{layers}\frac1N_\text{heads}\sum_\text{\it{layers}}\sum_\text{\it{heads}}\theta\left(g_i\cdot g_j\right)
\end{equation}

where $\theta(x)$ is the Heaviside step function. This is a scalar value that summarizes the box plot and we show the progression of alignment of the gradients as training progress and the end of the training (Table~\ref{tab:align} and Table~\ref{tab:grad_results} in Appendix D respectively). Note that, in the above formula, the token embedding layer is included in the computation and it is assumed to have $1$ head.

\begin{table}[h]
\centering
\resizebox{\columnwidth}{!}{
\begin{tabular}{@{}l l l l l l @{}}
\toprule
Dataset & Task Comparisons & Epoch 1 & Epoch 2 & Epoch 3 & Epoch 4 \\ \midrule
\multirow{6}{*}{Cervical} & Cord-Stenosis      & .46 & .67 & .75 & .81\\ \cmidrule(l){2-6} 
                          & Cord-Disc          & .37 & .52 & .69 &  .74\\ \cmidrule(l){2-6} 
                          & Cord-Foraminal     & .49 & .61 & .77 & .83\\ \cmidrule(l){2-6} 
                          & Disc-Stenosis      & .51 & .62 & .69 & .78\\ \cmidrule(l){2-6} 
                          & Disc-Foraminal     &  .47 & .59 & .65 & .73\\ \cmidrule(l){2-6} 
                          & Foraminal-Stenosis & .54 &  .66 & .72 & .79\\ \midrule
Lumbar                    & Disc-Stenosis      &   .44   &  .53 & .59 &  .68 \\ \cmidrule(l){2-6}
                          & Nerve-Stenosis     &   .51   & .57   & .66 & .73\\ \cmidrule(l){2-6} 
                          & Disc-Nerve         &    .48  & .55  &  .63 & .71 \\ \bottomrule
\end{tabular}
}
\caption{Results showing the Average Proportion of Aligned Gradients between various task specific models at various epochs.}
\label{tab:align}
\end{table}

To summarize: The task specific models not only show similar representations but they arrive at these representations by moving in a similar direction after starting from the pretrained weights. We would also like to point that we observe similar behavior when we run our experiments with the BERT~\citep{devlin2019bert} and the Clinical BERT~\citep{alsentzer2019publicly} models. 

\section{Results on Multi-Task Models}
In this section, we give empirical evidence on the success of MTL for our datasets. The results shown in this section are from our test set.

For our classification task, the PubMedBERT model is used as the backbone. This BERT model is finetuned on the the cervical tasks resulting in $4$ task-specific BERT sequence classifier models which provides our baseline results. For the lumbar dataset, the PubMedBERT model is finetuned on the $3$ classification tasks resulting in $3$ task-specific BERT sequence classifier models. 

Now, instead of finetuning the task specific models for extracting various pathology information from the cervical spine dataset, $4$ classifier heads (i.e. $4$ linear layers) are added to a single PubMedBERT model to create an output layer of shape $[3,3,2,2]$, where the first $3$ outputs correspond to the logits for the stenosis severity prediction, the next $3$ for the disc severity, the next $2$ for the cord severity and the final $2$ logits for the foraminal severity. For the lumbar dataset, $3$ classifier heads are added to the PubMedBERT model to create an output layer of shape $[3,3,2]$, where the first $3$ outputs correspond to the logits for the stenosis severity prediction, the next $3$ for the disc severity, and the final $2$ logits for the nerve severity.

For the experiments, with both the datasets, a dropout of $.5$ is added to the BERT vectors before passing them to the classifier layers. Each of these classifier heads is trained with a cross entropy loss with the predicted logits and the ground truth targets. All the losses are added up which allows the gradients to backpropagate through the whole model and train these classifier heads jointly. 

The results for our experiments are shown in Table~\ref{tab:Expt-results-lumbar} for the lumbar dataset and Table~\ref{tab:Expt-results-cervical} for the cervical dataset.
\begin{table}[h]
\resizebox{\columnwidth}{!}{
\begin{tabular}{llccc}
Backbone & Model & Disc & Stenosis & Nerve \\
\midrule
\begin{tabular}[c]{@{}c@{}} BERT \\ BASE \end{tabular}  & \begin{tabular}[c]{@{}c@{}}  Baseline\\ (single tasker) \end{tabular}  & $.78 \pm .03$ & $.79 \pm .02$ & $.8 \pm .03$  \\
 & Multi-Tasking & $.77 \pm .02$ & $.78 \pm .01$ & $.79 \pm .02$ \\

 \midrule 
\begin{tabular}[c]{@{}c@{}} CLINICAL \\ BERT \end{tabular} 
 & \begin{tabular}[c]{@{}c@{}}  Baseline\\ (single tasker) \end{tabular}  & $.81 \pm .03$ & $.83 \pm .02$ & $.82 \pm .03$  \\
 & Multi-Tasking & $.83 \pm .02$ & $.8 \pm .04$ & $.81 \pm .02$ \\
\midrule 
\begin{tabular}[c]{@{}c@{}} MSR PubMedBERT \end{tabular} 
 & \begin{tabular}[c]{@{}c@{}}  Baseline\\ (single tasker) \end{tabular}  & $.82 \pm .03$ & $.83 \pm .03$ & $.81 \pm .04$ \\
 & Multi-Tasking & $\mathbf{.84 \pm .01}$ & $\mathbf{.84 \pm .03}$ & $\mathbf{.86 \pm .04}$ \\

\bottomrule
\end{tabular}
}
\caption{Table showing the macro F1 scores over 5 trials of our Baseline and Multi-Tasking Models on the Lumbar Dataset.}
\label{tab:Expt-results-lumbar}
\end{table}

For fair comparisons, we also conduct experiments with the BERT base and the Clinical BERT models as well. We notice that the PubMedBERT produces slightly better results than both the Clinical BERT and the BERT base. We believe this is due to the fact that the vocabulary for PubMedBERT is tailored for clinical text, unlike that of Clinical BERT, which uses the same vocabulary as that of BERT. 

\begin{table}[h]
\resizebox{\columnwidth}{!}{
\begin{tabular}{llcccc}
Backbone & Model & Stenosis & Disc & Cord & Foraminal \\
\midrule
\begin{tabular}[c]{@{}c@{}} BERT \\ BASE \end{tabular}  & \begin{tabular}[c]{@{}c@{}}  Baseline\\ (single tasker) \end{tabular}  & $.62 \pm .03$ & $.64 \pm .03$ & $.70 \pm .03$ & $.79 \pm .03$ \\
 & Multi-Tasking & $.62 \pm .02$ & $.65 \pm .03$ & $.72 \pm .02$ & $.78 \pm .01 $\\

 \midrule 
\begin{tabular}[c]{@{}c@{}} CLINICAL \\ BERT \end{tabular} 
 & \begin{tabular}[c]{@{}c@{}}  Baseline\\ (single tasker) \end{tabular}  & $.64 \pm .05$ & $.66 \pm .02$ & $.71 \pm .02$ & $.82 \pm .01$ \\
 & Multi-Tasking & $.63 \pm .02$ & $.67 \pm .01$ & $.75 \pm .01$ & $.79 \pm .03 $\\
\midrule 
\begin{tabular}[c]{@{}c@{}} MSR PubMedBERT \end{tabular} 
 & \begin{tabular}[c]{@{}c@{}}  Baseline\\ (single tasker) \end{tabular}  & $.66 \pm .03$ & $.68 \pm .04$ & $\mathbf{.73 \pm .05}$ & $\mathbf{.84 \pm .01}$ \\
 & Multi-Tasking & $\mathbf{.67 \pm .01}$ & $\mathbf{.69 \pm .01}$ & $.72 \pm .04$ & $.83 \pm .03 $\\

\bottomrule
\end{tabular}
}
\caption{Table showing the macro F1 scores over 5 trials of our Baseline and  Multi-Tasking Models on the Cervical Dataset.}
\label{tab:Expt-results-cervical}
\end{table}

The hyperparameters and other training and implementation details can be found in Appendix C. 

\section{Deployment}
We deploy our spine pipeline system on an AWS p3.2x machine with a single NVIDIA V100 GPU. Reports are passed through the pipeline daily and first go through the report segmenter which tags sentences belonging to our set of motion segments. Post-processing is done per report to aggregate sentences belonging to each motion segment group and to filter out any reports that do not contain motion segments. Each grouping of motion segments is individually classified through our MTL model to predict a severity class per pathology. Both the report segmenter and the multi-tasking model are processed in batch mode with latencies of 31ms/report and 56ms/report, respectively. Compared to single pathology models, we observe a 3x improvement in latency per study when using the MTL pathology model. The spine pipeline is routinely evaluated in an offline setting for studies that do not produce any motion segment groupings or fail to capture any sentences for a given motion segment, per report. Our current deployment only supports the lumbar reports and we are in the process of extending our deployment to also support the cervical pathologies. 

\section{Conclusion and Future Work}
In this work, a simple multi-tasking model is presented that is competitive with task specific models. Instead of training and deploying task specific models, only one model is trained and deployed. This allows us to save significant costs during training and faster inference during deployment while achieving significant model compression, without any loss in the quality of performance. Our work opens the possibility of using multi-tasking models to extract information over various different body parts, allowing users to leverage large transformer models using limited compute resources. 

Our novel pipeline is one of the very few works that attempts to extract pathologies and their severities from a heterogeneous source of radiologists' notes on lumbar and cervical spine MRIs at the level of \textit{motion segments}. These findings suggest that our approach may not only be more widely generalizable and applicable, but also more clinically actionable.  

We believe our analysis with CKA and gradient alignment sheds more light on the success of MTL. This insight has led to our process change from single-task BERT based models to a more cost-effective MTL system.
Our analysis is widely applicable for other datasets and tasks. 

It is tempting to ask if one can use one multi-task model for both the lumbar and the cervical datasets. This is a work in progress and we have found strong similarity between single task models in the two datasets (most notably between the lumbar disc and the cervical disc models and the lumbar stenosis and the cervical stenosis models). However, unlike in the above analysis, we see low CKA scores between various other task specific models which may make MTL difficult (see Appendix D). We are in the process of using our analysis, along with insights borrowed from~\citep{standley2020tasks, yu2020gradient} to either group tasks from the two datasets or align different task-specific gradients to create an efficient learner. 

The biggest drawback of our work is the limited amount of data on which our observations are verified. We are actively addressing this issue as we annotate more reports concerning various pathologies in different body parts.

\section*{Ethical Considerations}
Because of legal and institutional concerns arising
from the sensitivity of clinical data, it is difficult for the NLP community to gain access to relevant data except for MIMIC~\citep{Johnson2016}. Despite its large size (covering over $58k$
hospital admissions), it is only representative of
patients from a particular clinical domain (the intensive care unit) and geographic location
(a single hospital in the United States). Such a sample is not representative of either larger population of patient admissions or other geographical regions/hospital systems. We have tried to address the second issue by collecting data across multiple practices in the US. However, it is impossible to predict whether our models will generalize to the entire patient population without actually evaluating on \textit{all} the different radiology practices. Thus we have to be extra careful about out-of-distribution data since the actionable insights we generate from our models can be potentially faulty and can lead to severe consequences. 

Finally, we recognize the need to minimize ethical risks of AI implementation which can include threats to privacy and confidentiality, informed consent, and patient autonomy. We strongly believe that stakeholders should be encouraged to be flexible in incorporating AI technology, most likely as a complementary tool and not a replacement for a physician. Thus, we develop our workflows, annotation guidelines and generate actionable insights by working in conjunction with a varied group of radiologists and medical professionals. 

\bibliography{anthology,naacl_modified}
\bibliographystyle{acl_natbib}

\appendix

\section{Example of our Dataset}
\begin{figure}[h]
\centering
\includegraphics[width=\columnwidth]{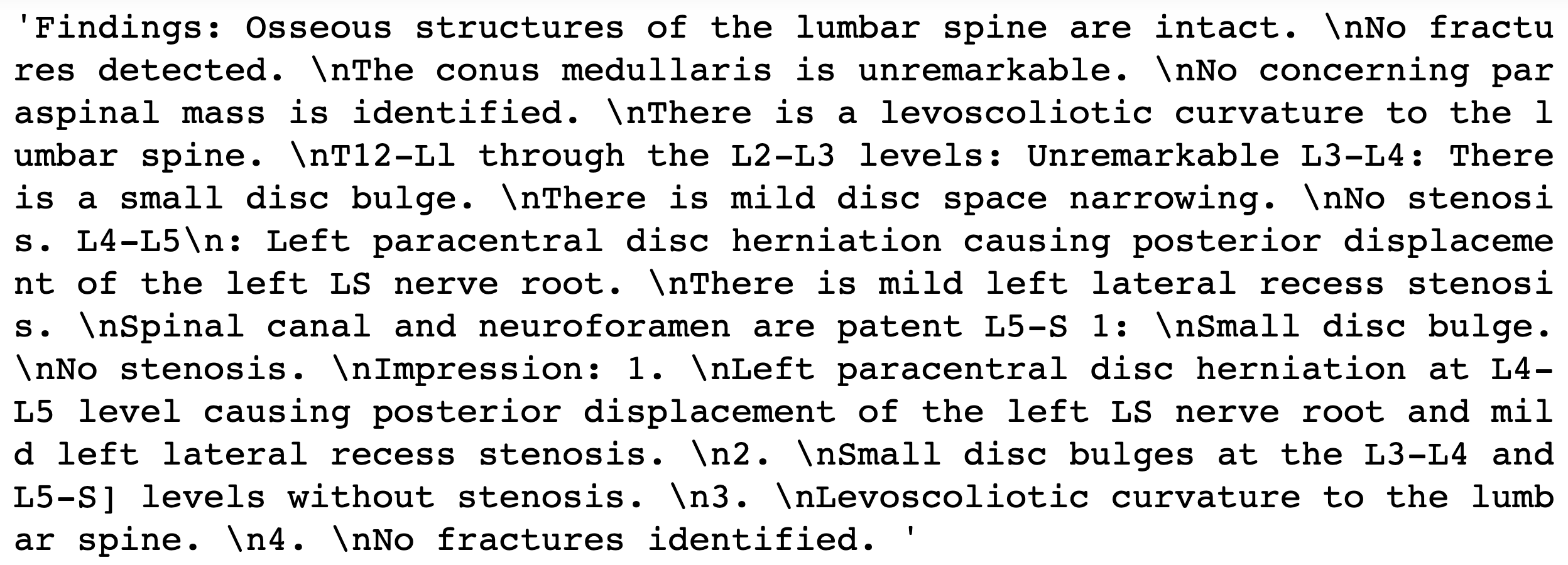}
\caption{An example of a report from our Lumbar Dataset. }
\label{fig:lumbar-dataset}
\end{figure} 
In this section, we will show some examples of lumbar and cervical reports from our dataset.
\begin{figure}[h]
\centering
\includegraphics[width=\columnwidth]{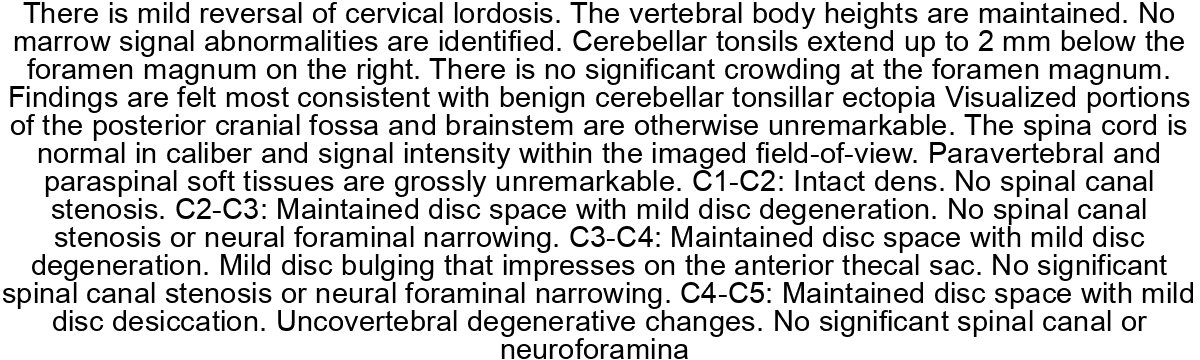}
\caption{An example of a report from our Cervical Dataset. }
\label{fig:cervical-dataset}
\end{figure} 

\begin{table*}[t]
\begin{center}
\resizebox{\textwidth}{!}
{
\begin{tabular}{l l l l l l}

Hyperparameter Type & \begin{tabular}[c]{@{}c@{}}  Single Tasking Models\\ on Cervical Dataset  \end{tabular} & \begin{tabular}[c]{@{}c@{}}  Multi-Tasking Models\\ on Cervical Dataset  \end{tabular} & \begin{tabular}[c]{@{}c@{}}  Single Tasking Models\\ on Lumbar Dataset  \end{tabular} & \begin{tabular}[c]{@{}c@{}}  Multi-Tasking Models\\ on Cervical Dataset  \end{tabular} & NER\\
\toprule
Epochs & 5 & 12 & 6 & 11 & 5\\
Batch Size & 16 & 16 & 16 & 16 & 16\\
Sequence Length & 512 & 512 & 512 & 512 & 256\\
Optimizer &  AdamW & AdamW & AdamW & AdamW & AdamW \\
Learning Rate & 2e-5 & 3e-5 & 2e-5 & 3e-5 & 1e-5\\
Weight Decay & 1e-4 & 1e-4 & 1e-4 & 1e-4 & 1e-3\\
Gradient Clip & 2 & 5 & 2 & 5 & 2\\
Early Stopping & Yes & Yes & Yes & Yes & Yes\\
Learning Rate Scheduler & Linear & Linear & Linear & Linear & Linear \\

\bottomrule

\end{tabular}
}
\caption{Hyperparameters used for all our experiments}
\label{table: Hyperparameters}
\end{center}
\end{table*}
\section{More Details about our Workflow}
In this section, we give a more detailed description of our novel workflow. Our main goal is to detect pathologies at the \textit{motion segment} level from radiologists' MRI reports.  The motion segments in the cervical reports that we are interested in are C2-C3, C3-C4, C4-C5, C5-C6, C6-C7 and C7-T1 and the motion segments of interest in the lumbar reports are L1-L2, L2-L3, L3-L4, L4-L5 and L5-S1. We first make sure that the reports are de-identified and then use a Spacy~\cite{spacy} parser to break the report into sentences. Then each sentence is tagged by annotators and they are given labels of various pathologies and their severities if the sentence mentions that pathology. To detect pathologies at a motion segment level, we use our BERT based NER system to tag the locations present in each sentence. Our BERT based NER model is a binary classifier model (Location Tag vs the Other Tag). It is is trained on both lumbar and cervical MRI reports that can predict the location tags in those reports. Our NER model achieves an F1 score of $.9$. 

We then use an appropriate body part specific rule based system to group all sentences to the correct motion segment. If a sentence does not explicitly have a motion segment mentioned in it, we use a rule based method to assign the sentence to one of the above mentioned motion segments or to a generic category ''No motion segments found". Given the disparate source of our data and due to typos and OCR errors, for example, L23, L2L3, L@L3, $L2\_L3$ all may refer to the motion segment L2-L3 and thus our systems are mindful of this diversity of the clinical notes. Finally to use our BERT based models for pathology detection on the level of motion segments for a given report, we concatenate all sentences for a given motion segment and use the [CLS] token for the segment that is used for the downstream classification task.  

Since we are interested in predictions at the motion segment level, we do not use the sentences that are grouped under ''No motion segments found" to train the classifier models, nor do we evaluate our classifier models on those sentences.

\section{Hyperparameters and Other Training Details}
We create a validation set using $20\%$ of the samples of the training set where the samples are drawn via stratified samples so the data distribution is maintained across splits. The hyperparameters used for training the NER model and various classification models can be found in Table~\ref{table: Hyperparameters}.

PyTorch~\citep{NEURIPS2019pytorch} and the HuggingFace library~\citep{wolfetal2020transformers} is used to conduct our experiments which are run on 1 NVIDIA V100 16GB GPU.

\begin{figure*}[h]
\centering
\includegraphics[width=16cm]{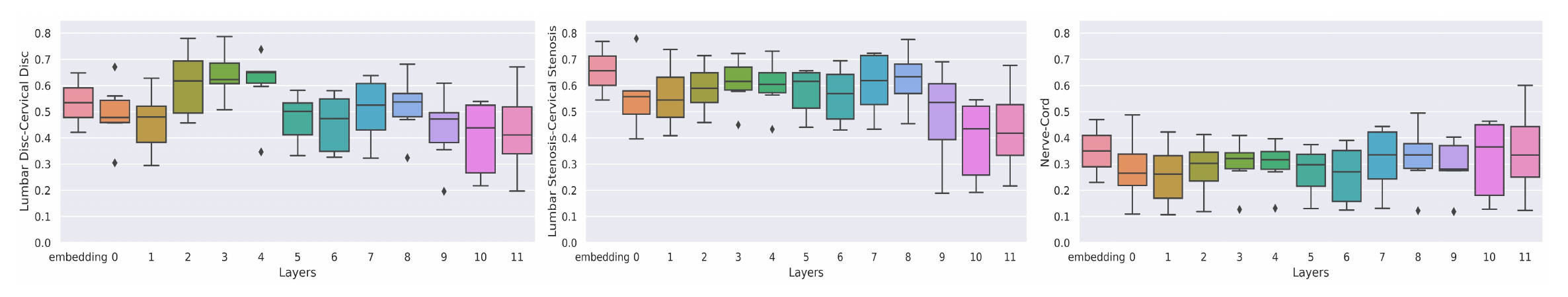}
\caption{Box plot showing CKA scores between models trained on tasks on the lumbar and the cervical dataset. The y-axis is chosen to be (0,.85). The figure shows the low CKA scores between the cord and the nerve models and high scores between the stenosis models and the disc models.}
\label{fig:intra-cka}
\end{figure*}

\section{Additional CKA Results and Gradient Alignment Results}
In this section, we present some additional results on comparing representations between our various models. 

We present the average proportion of aligned gradients (APAG) at the end of training in Table~\ref{tab:grad_results}.
We also compute the cosine similarity between the gradients. We then take the average of them for a given layer, thus yielding a scalar value per layer. This yields cosine similarity values which are over 90 \% positive. For simplicity, we average those numbers to produce a scalar value that measures the cosine similarity between the gradients of two models. Model level statistics can be found in Table~\ref{tab:grad_results}.
\begin{table}[h]
\centering
\resizebox{\columnwidth}{!}{
\begin{tabular}{@{}l l l l @{}}
\toprule
Dataset & Task Comparisons & Cosine Similarity & \begin{tabular}[c]{@{}l@{}}Average Proportion \\ of Aligned Gradients\end{tabular} \\ \midrule
\multirow{6}{*}{Cervical} & Cord-Stenosis      & .013 & .89 \\ \cmidrule(l){2-4} 
                          & Cord-Disc          & .005 & .87 \\ \cmidrule(l){2-4} 
                          & Cord-Foraminal     & .011 & .91 \\ \cmidrule(l){2-4} 
                          & Disc-Stenosis      & .012 & .88 \\ \cmidrule(l){2-4} 
                          & Disc-Foraminal     & .007 & .87 \\ \cmidrule(l){2-4} 
                          & Foraminal-Stenosis & .005 & .84 \\ \midrule
Lumbar                    & Disc-Stenosis      &  .008    &  .75   \\ \cmidrule(l){2-4}
                          & Nerve-Stenosis     &   .01   &  .86   \\ \cmidrule(l){2-4} 
                          & Disc-Nerve         &   .002   &    .86 \\ \bottomrule
\end{tabular}
}
\caption{Results showing the Cosine Similarity and the Average Proportion of Aligned Gradients between various task specific models after the end of training.}
\label{tab:grad_results}
\end{table}

Given some similarities between certain label spaces in the lumbar and the cervical dataset (particularly for the disc herniation and the central canal stenosis labels), we believe that some task specific models between tasks across datasets may show similar representations. To validate this hypothesis, we computed the CKA between lumbar stenosis and the cervical stenosis models and the lumbar disc and the cervical disc models. The natural question is : what happens to the single tasking models that are trained on label spaces that are semantically different? Fig~\ref{fig:intra-cka} shows low CKA scores between the cord and the nerve models. This is an active work in progress to be able to group similar tasks~\citep{standley2020tasks} to create a MTL framework that works for both the cervical and the lumbar spine. Another future direction is to use realign gradients using the techniques in~\citep{yu2020gradient}. However to realign the gradients, one has to save the entire computation graph after the backward pass via \textit{loss.backward(retain\_graph=True)} which becomes a bottleneck for large transformer models. To mitigate this issue, one can use parameter efficient methods like adapters which we have shown to work in these MTL settings in our previous work~\citep{cervicalabstract}.

\section{Annotation Process}
All data are annotated by our team of inhouse annotators with clinical expertise. All annotators are trained for the given task and provided clear guidelines on the task and performance is measured periodically on a benchmark set and feedback is provided. 

\end{document}